\pdfoutput=1
\documentclass{article}

   \PassOptionsToPackage{numbers}{natbib}

\usepackage[final]{neurips_2024}

\usepackage[utf8]{inputenc} 
\usepackage[T1]{fontenc}    
\usepackage{hyperref}       
\usepackage{url}            
\usepackage{booktabs}       
\usepackage{amsfonts}       
\usepackage{nicefrac}       
\usepackage{microtype}      
\usepackage{xcolor}         
\usepackage{amsmath}
\usepackage{booktabs}
\usepackage{graphicx}
\usepackage{xcolor}

\title{From Context to Action: Analysis of the Impact of State Representation and Context on the Generalization of Multi-Turn Web Navigation Agents}

\author{
    Nalin Tiwary${^*}$\textsuperscript{\textdagger}
    \And
    Vardhan Dongre${^*}$\textsuperscript{\textdagger}
    \And
    Sanil Arun Chawla${^*}$
    \AND
    Ashwin Lamani${^*}$
    \And
    Dilek Hakkani-Tür${^*}$
}

\begin{document}
\maketitle



\begin{abstract} 
\renewcommand{\thefootnote}{\fnsymbol{footnote}}
   \footnotetext[2]{Equal Contribution}
    \footnotetext[1]{Department of Computer Science, University of Illinois-Urbana Champaign}
Recent advancements in Large Language Model (LLM)-based frameworks have extended their capabilities to complex real-world applications, such as interactive web navigation. These systems, driven by user commands, navigate web browsers to complete tasks through multi-turn dialogues, offering both innovative opportunities and significant challenges. Despite the introduction of benchmarks for conversational web navigation, a detailed understanding of the key contextual components that influence the performance of these agents remains elusive. This study aims to fill this gap by analyzing the various contextual elements crucial to the functioning of web navigation agents. We investigate the optimization of context management, focusing on the influence of interaction history and web page representation. Our work highlights improved agent performance across out-of-distribution scenarios, including unseen websites, categories, and geographic locations through effective context management. These findings provide insights into the design and optimization of LLM-based agents, enabling more accurate and effective web navigation in real-world applications.
\end{abstract}

\section{Introduction} \label{sec:Introduction}




Conversational Web Agents (CWAs) have emerged as a promising way of executing complex tasks within web browsers by engaging in multi-turn dialogues with users. These agents are designed to navigate web pages and complete tasks based on user instructions, offering a blend of interaction and automation. While most existing work focuses on single-turn navigation, multi-turn navigation is crucial because it mirrors the complexity and dynamic nature of real-world web interactions, allowing for more human-like, adaptable, and error-resilient communication. Recently introduced benchmarks such as WebLINX \citet{pmlr-v235-lu24e} and MT-Mind2Web \cite{deng2024multi} reflect the pivot towards multi-turn navigation.

\begin{figure*}[ht]
    \centering
    \includegraphics[width=\linewidth]{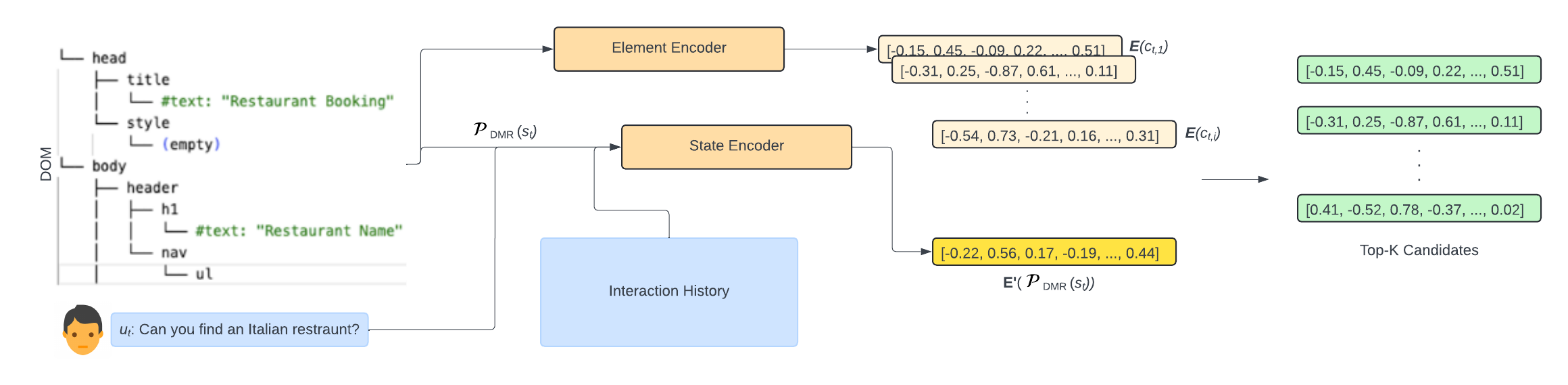}
    \caption{Dense Markup Ranking:  Figure illustrates the DMR process, where a web page's Document Object Model (DOM) is generated from the HTML and is parsed, and both elements and the current state are encoded into vectors. We compute the cosine similarity for each candidate element and rank the elements relevant to the user's query to facilitate informed navigation and interaction decisions.}
    \label{fig:dmr}
\end{figure*}


A critical challenge remains in understanding how the context impacts the performance and generalization capabilities of CWAs. To avoid giving the long and noisy HTML page directly to an agent and instructing it to predict the correct action, the CWAs introduced in these works \cite{pmlr-v235-lu24e, deng2024multi} approach web navigation in a two-stage process. First, they identify the most relevant elements from the Document Object Model (DOM)\footnote{Tree representation of the HTML page} Tree of the webpage. These elements are combined with interaction history and screenshots (in a multi-modal context) to construct an input representation for the LLM. In the second stage, the agents predict the appropriate actions to take based on this comprehensive input. In WebLINX, this two-stage process leverages compact text-retrieval architectures based on a dual encoder-based approach called Dense Markup Ranking (DMR) to pair DOM elements with task elements(pictured in Fig \ref{fig:dmr}), which in turn reduces the input to the action model from the entire web page to a few candidate HTML elements. The priority of effectively managing context is crucial because the richness and relevance of the input directly dictates the agent’s ability to make accurate decisions. These works reflect this priority through processes like truncating the DOM tree or limiting interaction history. 

\begin{figure*}[ht]
    \centering
    \includegraphics[width=0.8\linewidth]{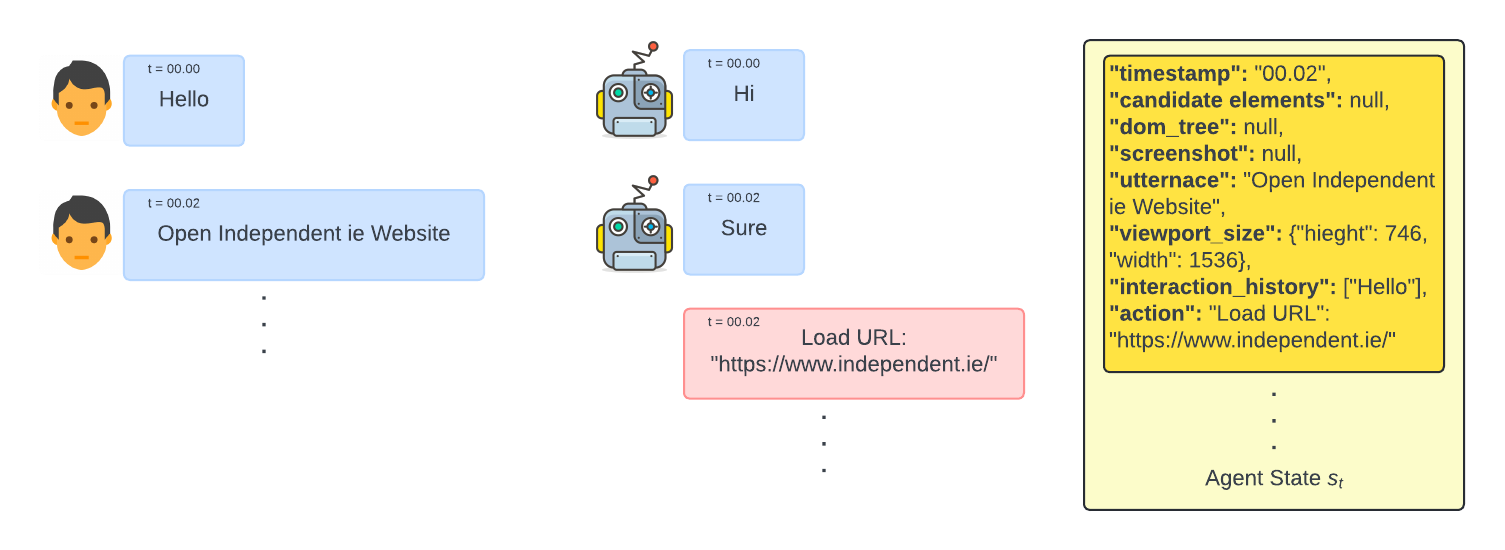}
    \caption{The figure depicts the interaction between a user and a web agent, illustrating each turn. The user's and agent's utterances are displayed in blue speech bubbles on the left and right, while the agent's actions are shown in red on the right. The red box shows a single agent action along with the arguments associated with each action, such as the URL to load. The agent's internal state, including its understanding and actions at each point (timestamped as "t"), is highlighted in yellow. }
    \label{fig:web agent full}
\end{figure*}

Our work aims to provide a systematic understanding of the contextual elements that influence the performance of CWAs by analyzing the impact of state representation and judicious context management on the generalization of multi-turn web navigation agents. This, in turn, allows us to pinpoint the essential components of web navigation that serve as better contexts, allowing agents to accumulate and leverage information from ongoing dialogues. This enables a deeper comprehension of user intentions over the course of an interaction. We show how careful management of factors, such as interaction history and truncation of the agent's state(pictured in Fig \ref{fig:web agent full}), can improve web navigation performance. We found that increasing interaction history from the last 5 to the last 10 turns, along with providing the whole DOM tree in the context, improves recall in the markup ranking process by $4.37\%$ across out-of-distribution (OOD) scenarios—such as previously unseen websites, categories, and web domains from unseen geographic locations.

\section{Related Work} \label{sec:Related Work}

\subsection{Web Navigation Agents} \label{sub:Web Navigation Agents}
Web navigation agents have become increasingly sophisticated, evolving from simple rule-based systems to complex agents capable of completing real-world tasks within web browsers. Early work in this area focused on single-task web agents \cite{zheran2018reinforcement}, on datasets like MiniWoB++ \cite{pmlr-v70-shi17a}, which employed task-specific primitives to simulate web navigation in controlled environments. More recently, environments like WebShop \cite{yao2022webshop}, containing human-written task instructions for e-commerce settings, and WebArena \cite{zhou2023webarena}, containing task instruction across e-commerce, forum discussions, software development, and content management, have expanded the scope to more diverse and realistic web interactions. These works, along with works such as MIND2WEB \cite{deng2023mindweb}, have facilitated resources for building  SeeAct \cite{zheng2024gptvision} have introduced resources building autonomous navigation agents like SeeAct \cite{zheng2024gptvision}. These frameworks, while robust, often emphasize structured navigation tasks with limited complexity in dialogue or context management. They lack the diversity and complexity brought on by multi-turn conversational dialogues. Additionally, there is limited focus on the amount and intricacy of context or content length that LLMs can and should utilize to fulfill these navigation tasks. This is a crucial aspect, as real-world interactions often require agents to understand and remember details from extended dialogues to effectively respond to user inquiries and commands over time.

\subsection{Website Representation} \label{sub:Website Representation}
Representing real-world websites is a complex task due to HTML pages' extensive and dynamic nature. Prior approaches to simplifying or compressing the textual representation of websites include rule-based algorithms \cite{zhou2021simplified}, graph embeddings \cite{wang2022webformer}, and model-based approaches \cite{deng2022dom}. These methods aim to make the content of web pages more manageable for models to process. For this work, we worked with DOM elements to make suitable comparisons with dense markup retrievers introduced in \cite{pmlr-v235-lu24e}.

\section{Methods} \label{sec:Methods}

In this section, we discuss the two-stage process for web navigation based on appropriate candidate selection, followed by selecting action(s) based on the instructor's commands. In each interaction, at each turn, the agent's state $s_t$ includes the following, based on their availability, to predict actions $a_t$:
\begin{itemize}
    \item $\boldsymbol{c_t}$: Candidate elements from DOM that are possible targets for actions during a web navigation task
    \item $\boldsymbol{d_t}$: Current DOM tree of webpage
    \item $\boldsymbol{i_t}$: Screenshot of web browser
    \item $\boldsymbol{u_t}$: Instructor's utterance 
    \item $\boldsymbol{h_t}$: Interaction History
\end{itemize}

\subsection{Dense Markup Ranking (DMR)} \label{sub:DMR}
Web pages are typically represented by large and intricate Document Object Model (DOM) trees, making direct processing by large language models (LLMs) computationally infeasible. The goal of the DMR stage, as in \cite{pmlr-v235-lu24e}, is to efficiently prune and rank HTML elements, identifying those most relevant to the user’s task and instructions based on the utterance and interaction history.

Given an HTML document represented as a DOM tree with numerous elements, the DMR aims to rank these elements based on their relevance to the task defined by user instructions and the interaction history. We adopt a dual encoder-based approach as described in \cite{pmlr-v235-lu24e}. Each HTML element is represented using its textual content and attributes, which are then encoded along with user utterances and past interaction history using separate encoders. Using a cosine similarity-based objective, we learn to rank elements based on their relevance. The process begins by gathering the current state of the agent, which includes current DOM tree $(d_t)$, user utterance $(u_t)$, and past interaction history $(h_t)$.

 For each turn \( t \), we have the processed text representation of the state \( \mathcal{P}_{\text{DMR}}(s_t) \), which is used to score candidate elements \( c_{t,i} \),  where \( c_{t,i} \) is the $i^{th}$ candidate element among the many that might be considered at a given turn. The candidate element \( c_{t,i} \) is also represented as text. The 2 Encoders \( E(x) \) and \( E'(x) \) encodes this input text to output vectors. 

\begin{itemize}
    \item If \( c_{t,i} \) is the target candidate, we set the label \( y(c_{t,i}) = 1 \).
    \item Otherwise, \( y(c_{t,i}) = 0 \).
\end{itemize}

The cosine similarity loss is defined as the following mean-squared error:

\[
\mathcal{L}_t = \left\| y(c_{t,i}) - \text{sim}_{\cos}\left( E'(\mathcal{P}_{\text{DMR}}(s_t)), E(c_{t,i}) \right) \right\|_2^2
\]

where the cosine similarity is defined as:

\[
\text{sim}_{\cos}(x, y) = \frac{x \cdot y}{\|x\| \|y\|}
\]

During inference, the cosine similarity is used to generate a score for each candidate representing the similarity between \( \mathcal{P}_{\text{DMR}}(s_t) \) and candidate \( c_{t,i} \) at turn \( t \). The score is used to rank the candidates and choose the top-\( k \) candidates for the action prediction stage.

\subsection{Action Model} \label{sub:Action}
Once the DMR identifies the relevant DOM elements, these candidates are combined with other information from the state $s_t$ to construct a representation for predicting action strings, which can be parsed and executed. The Action model integrates multiple input modalities, including text instructions, screenshots of the web page, and the interaction history. The input to the model is truncated strategically to fit within the model's token limit. The actions include commands like \textbf{click}, \textbf{text\_input}, and \textbf{submit}, which are necessary for navigating and interacting with web pages. \cite{pmlr-v235-lu24e} designed a hierarchical truncation process by setting a limit for each component (DOM tree, utterances, actions, candidates, viewport size, screenshot of the browser). At each turn $t$, the action model takes the constructed input $x_t$ and uses it to predict the action $a_t$

\section{Experiments} \label{sec:Experiments}

\subsection{Dataset} \label{sub:Dataset}

Our candidate dataset for this study is the WebLINX dataset, comprised of demonstrations collected during real-time interaction between a commander and a human navigator. It includes 100,000 interactions across 2,337 demonstrations of conversational web navigation, all produced by human experts. The data spans 155 real-world websites, carefully selected from 15 different geographic areas, providing a broad and diverse context for our analysis. The \textbf{Train} split contains 969 demos, which we use for fine-tuning our models, and 1268 demos for testing. Table \ref{tab:datasets} provides insight into the makeup of a demonstration within the dataset. In Table \ref{tab:datasets}, the utterance length is measured by the number of characters, the duration of the demo is measured by the number of seconds elapsed between the first instructor utterance and the last turn in the demo, and the number of active turns is the number of turns from the split used in the training or evaluation process.

\begin{table}[ht]
\centering
{
\caption{Comparing demonstration in the Train vs Test-OOD split\\}
\resizebox{0.5\columnwidth}{!}{
\begin{tabular}{ cc c} \toprule
        \textbf{Metric}   & \textbf{Train} & \textbf{Test-OOD}  \\ \midrule
Avg. Number of Turns & 44.93 &  41.61 \\ 
Avg. Number of Instructor Turns & 9.22 & 7.64 \\ 
Avg. Number of Navigator Turns & 7.97 & 8.82 \\ 
Avg. Instructor Utterance Length & 44.74 & 46.38 \\ 
Avg. Navigator Utterance Length &  61.77 & 66.17 \\ 
Avg duration of a demo & 398.27s & 473.18s  \\
Number of Active Turns & 24118 & 23029 \\ \bottomrule
\end{tabular}
}
\label{tab:datasets}
}

\end{table}

We specifically considered the OOD samples available in the dataset for our evaluations to assess the impact on generalizability. The \textbf{Test-OOD} set contains 1168 demos and several distinct OOD test subsets to evaluate web navigation agents across multiple scenarios. 211 \textbf{Test-Web} demos for testing the agents' ability to handle completely unseen websites within recognized subcategories, while 223 \textbf{Test-Cat} demos introduces entirely new subcategories within the 8 known categories. 290 \textbf{Test-Geo} demos assesses how well the agents perform in geographic locations not included in the training set. Lastly, 444 \textbf{Test-Vis} demos uniquely tests the agents' performance without visual feedback, simulating scenarios where visual cues are absent, pushing the limits of their navigation capabilities. Together, these test sets ensure a comprehensive assessment of the agents' adaptability and robustness in realistic web navigation tasks.

\subsection{Metrics} \label{sub:Metrics}

\subsubsection{DMR}

In our work, we use recall rates, which are standard metrics in information retrieval tasks such as text retrieval, to evaluate the performance of our DMR model. Recall rates measure the ability of the model to retrieve relevant items from a given set of candidates, making them well-suited for assessing the DMR model's candidate generation capabilities.

\begin{itemize}
        

    

    \item \textbf{Recall @ K} is used to evaluate our DMR model. This indicates how well the model performs in retrieving relevant items within the top K results of a query. We calculate Recall @ 1, 5, 10, 20, and 200 as a metric for evaluating the candidate generation from the DMR model. We report Recall@10 specifically for comparisons across different experimental settings and models due to its usage to rank DMR models on the WebLINX leaderboard. We also report Recall@1 since it presents a view of the basic expected success.
    
   \begin{align}
    \text{Recall@K} &= \frac{\text{true Positives}}{\text{true Positives} + \text{false Negatives}} \nonumber \\
                    &= \frac{p_p}{p_p + p_n}
    \end{align}
    
    \end{itemize}

\subsubsection{Action model} To showcase the downstream impact of a better DMR model through generating better candidates, we also evaluate the effect of the improved candidates and effective context management on an action model. We inherit the overall score metrics to evaluate our action  model from \cite{pmlr-v235-lu24e}

\begin{itemize}
    \item \textbf{Intent Match (IM)} assesses whether the model's predicted action intent aligns with the reference action's intent. Defined as $IM(a', a) = 1$ if the intents of the predicted action $a'$ and the reference action $a$ are identical, and $IM(a', a) = 0$ otherwise. This metric evaluates the model's ability to identify the intended action correctly. It is defined as:
    \[ IM(a', a) = \begin{cases} 
    1 & \text{if } \text{intent of } a' = \text{intent of } a \\
    0 & \text{otherwise}
    \end{cases} \]

    \item \textbf{Text Similarity (F1)} utilizes the F1-score based on character n-gram matches, denoted as \textbf{chrF}$(a', a)$ \cite{popovic2015chrf}, to measure the lexical similarity between the predicted and reference text arguments in actions. The F1-score based on character n-gram matches for textual content, adjusted by intent match:
    \[ \text{chrF} = IM(a', a) \times \frac{2 \cdot \text{Precision} \cdot \text{Recall}}{\text{Precision} + \text{Recall}} \]
    where Precision and Recall are derived from character n-gram overlaps between the predicted and reference texts.
    
    \item \textbf{Overall Score} combines all of the above metrics as a micro-average of turn-level scores across all actions to provide a comprehensive performance metric:
    \[ \text{Overall Score} = \frac{1}{N} \sum_{i=1}^{N} \text{score}(a'_i, a_i) \]
    where \(N\) is the total number of actions, and \(\text{score}(a'_i, a_i)\) is the calculated score for each action pair.
\end{itemize}

\subsection{Analyzing Importance of Context in Web Navigation} \label{sub:Mods}
In our study, we perform experiments with the DMR method to optimize its efficiency and effectiveness on real-world websites. These improvements are focused on redefining the truncation process and understanding the impact of interaction history in context.

\subsubsection{Impact of Interaction History}
The input interaction history is a combination of past user utterances and actions. In the WebLINX benchmark, the interaction history is essential for understanding the context of the instructor's utterance. Typically, the models in this benchmark are configured to process interaction histories limited to the past five interactions. Thus, it is crucial to investigate the influence of interaction history length to answer how far back the interaction must be rolled out for the models to perform well. We experiment with different lengths of past interactions, assuming that extending the history provides a deeper contextual understanding, enabling the model to predict user intentions better. However, extending the history too much might lead to context dilution and lead to the model emphasizing irrelevant information.

\subsubsection{Impact of Representation Lengths and Truncation}
As the Methods section mentions, each candidate element is represented by a text description, which is then tokenized for analysis. The token size for these descriptions can vary significantly, potentially influencing the model's ability to interpret and rank these elements accurately. The representation of these candidate elements has been constrained to 200 tokens in the original implementations of DMR. This raises pertinent questions about the impact of token size on the selection of target candidates, especially in scenarios involving complex or detailed element descriptions that might exceed this token limit. We perform experiments to investigate the effects of varying token sizes on the performance of the DMR process. By altering the token limit and evaluating the resulting performance changes, we aim to detect if an optimal tokenization length exists for representing candidate elements.  

Also, as specified in the Methods section, the input sequence for the action model is truncated to avoid exceeding the input limits allowed by the models. \cite{pmlr-v235-lu24e} perform truncation by decomposing the components of the input state into their sub-components. The entire DOM "$d_t$" includes the values and text content; for candidates "$c_t$" it involves the XPath and children tags; and for interactions, each utterance and action is considered a sub-component. These components are then rendered, tokenized, and truncated (where the truncation of each component is thresholded to avoid heavy penalization of shorter components) until the goal limit is reached.


\subsection{Models} \label{sub:Models}
Our ranking process is similar to one proposed in the WebLINX benchmark, where we model it as a text retrieval challenge. This involves using a model to encode a query and generate scores for ranking potential candidates. Given this context, we review various models originally developed for text retrieval tasks, as their transferability to similar tasks is often high. Specifically, we chose all-MiniLM-L6-v2, a model developed by \cite{reimers2019sentence} based on the MiniLM model \cite{wang2020minilm}, as well as two other compact models: bge-small-en-v1.5 (BGE) \cite{xiao2023c} and gte-base (GTE) \cite{li2023towards}

\section{Results} \label{sec:Results}

\begin{table}[ht]
\centering
\caption{Comparison of Recall@1,@5, \& @10 scores for the MiniLM model across four token length limits: 100, 200, 400 tokens, and no token limit, i.e., maximum token length for the model.\\}
\resizebox{0.5\columnwidth}{!}{
\begin{tabular}{cccc} \toprule
\textbf{Token Limit}                   & \textbf{Recall@1} & \textbf{Recall@5} & \textbf{Recall@10} \\ \midrule
100 & 18.81 & 42.8 & 52.37 \\ 
200 & 18.49 & 42.8 & 52.55 \\ 
400 & \textbf{18.97} & 43.06 & 52.19 \\ 
No limit & 18.86 & \textbf{43.12} & \textbf{52.9} \\ \bottomrule
\end{tabular}
}\\

\label{lengths}
\end{table}


Our results indicate that optimizing context management and DOM element representation improves the performance of DMR models in candidate selection and, subsequently, web navigation agents.

First, we investigate the effect of truncating the representation of DOM element candidates to fit various token limits in Table \ref{lengths}. Utilizing the MiniLM model, increasing the token length from 100 to the maximum allowed length generally improved the average Recall@1, @5, \& @10 scores on the OOD test splits. The highest performance was achieved using the maximum token length setting, with an average Recall@10 score of 52.9. This result suggests that providing more comprehensive text representations of DOM elements can enhance the model's ability to accurately rank and retrieve relevant candidates.

\begin{figure}[ht]
    \centering
    \includegraphics[width=0.6\linewidth]{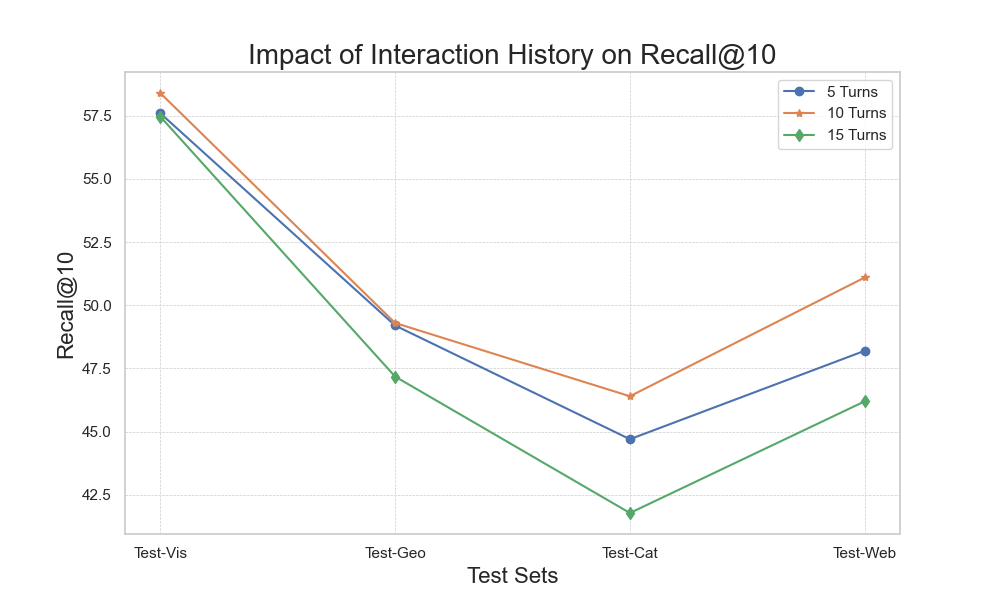}
    \caption{Recall@10 performance using MiniLM across interaction history lengths of 5,10, and 15 turns on the four Test-OOD splits. The plot highlights the importance of balancing a longer interaction history to enhance the model's ability to select relevant candidates in different test scenarios.}
    \label{fig:inthist}
    
\end{figure}

Figure \ref{fig:inthist} highlights explicitly the impact of the interaction history length on performance. The line plot demonstrates that extending the history context from 5 to 10 turns results in higher Recall@10 scores across all Test-OOD splits. This finding underscores the importance of providing a longer interaction context to the model, enabling it to understand user intentions better and select relevant DOM elements. However, the plot also reveals an interesting trend – extending the history further to 15 turns leads to a performance drop across the board. This suggests that while longer interaction contexts are generally beneficial, an excessively long history may introduce noise and adversely impact performance.

\begin{table*}[ht]
\centering
\caption{Comparison of Recall@10 scores for MiniLM, BGE, and GTE on the Test-OOD split.}
\resizebox{0.7\columnwidth}{!}{
\begin{tabular}{lcccccc} \toprule
    & \multicolumn{2}{c}{\textbf{MiniLM}} & \multicolumn{2}{c}{\textbf{BGE}} & \multicolumn{2}{c}{\textbf{GTE}} \\ \cmidrule(lr){2-3} \cmidrule(lr){4-5} \cmidrule(lr){6-7}
    & \textbf{Baseline} & \textbf{Ours} & \textbf{Baseline} & \textbf{Ours} & \textbf{Baseline} & \textbf{Ours} \\ \midrule
    \textbf{Test-OOD} & 51.87 & \textbf{54.13} & 50.01 & \textbf{50.31} & 48.16 & \textbf{50.34} \\ 
    \bottomrule
\end{tabular}
}

\label{dmr_comparison}
\end{table*}

\begin{table}[ht]
\centering
\caption{Recall@10 performance on Test-OOD splits for "our" MiniLM vs RoBERTA vs MiniLM}
\resizebox{0.5\columnwidth}{!}{
\begin{tabular}{cccc} \toprule
\textbf{Test Split}                   & \textbf{MiniLM(Baseline)} & \textbf{RoBERTA} & \textbf{MiniLM(Ours)} \\ \midrule
\textbf{Test-Web} & 52.75 & 54.65 & \textbf{54.96} \\ 
\textbf{Test-Cat} & 44.05 & \textbf{48.43} & 47.49 \\ 
\textbf{Test-Geo} & 50.95 & 52.76 & \textbf{53.98} \\ 
\textbf{Test-Vis} & 59.73 & \textbf{63.28} & 61.85 \\
\midrule
\textbf{Test-OOD} & 51.87 & \textbf{54.78} & 54.13 \\ \bottomrule
\end{tabular}
}\\

\label{dmr_compare}
\end{table}

\begin{table}[ht]
\centering
\caption{ Performance of "our"  MiniLM through Recall @1,@5, and @10 across Test-OOD splits.}
\resizebox{0.5\columnwidth}{!}{
\begin{tabular}{cccc} \toprule
\textbf{Test Split}                   & \textbf{Recall@1} & \textbf{Recall@5} & \textbf{Recall@10} \\ \midrule
\textbf{Test-Web} & 19.02 & 44.93 & 53.49 \\ 
\textbf{Test-Cat} & 17.11 & 38.9 & 46.93 \\ 
\textbf{Test-Geo} & 17.94 & 41.24 & 51.4 \\ 
\textbf{Test-Vis} & 21.37 & 47.41 & 59.78 \\
\midrule
\textbf{Test-OOD} & 18.86 & 43.12 & 52.9 \\ \bottomrule
\end{tabular}
}\\
\label{minilm_splits}
\end{table}

Utilizing the above experiments, we compared the performance of three models (MiniLM, BGE, and GTE) on the Test-OOD split of the WebLINX benchmark in Table \ref{dmr_comparison}. The baselines are the scores obtained from the WebLINX leaderboard. \footnote{https://mcgill-nlp.github.io/weblinx/leaderboard/} At the same time, "Ours" shows the results with our modifications of increasing the interaction history context to the last 10 turns and relaxing token length constraints for element representations. All three models showcase improvements over the baseline results; specifically, the best-performing DMR model in the baseline results, MiniLM, continues to be the best-performing model with an improvement of $4.37\%$.

With the increase in recall, MiniLM can rival much bigger candidate selection models, such as MindAct-RoBERTa, the cross-encoder proposed by \cite{deng2023mindweb}, which has a Recall@10 of $54.78$ on the Test-OOD split as reported in the appendix of \cite{pmlr-v235-lu24e}. Since our implementation of MiniLM preserves its nearly $5$x faster inference time  \footnote{See Appendix B.4 in \cite{pmlr-v235-lu24e}} compared to RoBERTa, the best DMR model in this comparison is clear.

Utilizing the candidates generated by "Our" MiniLM and an expanded interaction history from 5 to 10 turns, we trained the Flan-T5-250M \cite{chung2024scaling} to act as our action model. The average overall score on the Test-OOD split for the baseline from the WebLINX leaderboard is $14.99$. With our modifications, the overall score rose to $16.16$, highlighting that effective context management leads to an improved DMR, which downstream leads to better agent performance.


\section{Discussion} \label{sec:Discussion}

The experimental results presented in this work shed light on several key factors that influence the ability of web navigation agents to generalize effectively to out-of-distribution scenarios. Our modifications to the DMR improved its generalizability in real-time web navigation tasks. By systematically investigating the effects of interaction history length, token length constraints, and context representation, we uncover valuable insights about the factors influencing the web navigation task in multi-turn dialog settings and how to model these agents for better performance on unseen websites, categories, and geographic locations. Furthermore, by focusing on critical components, expanding the historical context, and unrestricted element representations, the model also gained performance on out-of-distribution samples.  



\section{Limitations and Future Work} \label{sec:Limitations and Future Work}
While our work aims to provide valuable insights into the impact of state representation and context management on the generalization of multi-turn web navigation agents, there are certain limitations in our work, which we aim to address as part of our future work. As part of future work, we aim to investigate the use of mechanisms that can dynamically determine the appropriate context boundary based on the specific out-of-distribution case. Such mechanisms, potentially leveraging contextualized pruning or attention techniques, can help identify and prioritize the most relevant dialogue segments and element descriptions, facilitating effective generalization while mitigating noise from excessive context. Additionally, we plan to study factors that influence the subsequent action selection process. We can further enhance these models' decision-making capabilities in out-of-distribution scenarios by understanding how different contextual elements and representation choices impact an agent's ability to predict appropriate actions based on the retrieved candidates.

\section{Conclusion} \label{sec:Conclusion}
Collectively, these findings emphasize the importance of effective context management and representation for enabling web navigation agents to generalize successfully to unseen domains. Our findings highlight the need to strike a balance for effective context management. While comprehensive contexts that empower a deeper understanding of user intent are advantageous, overly long histories could inadvertently introduce irrelevant information, hindering generalization. Judiciously determining the optimal context length and representation granularity is crucial for enabling robust performance across diverse, unseen scenarios.

\bibliographystyle{plainnat}
\bibliography{paper}

\begin{thebibliography}{17}
\providecommand{\natexlab}[1]{#1}
\providecommand{\url}[1]{\texttt{#1}}
\expandafter\ifx\csname urlstyle\endcsname\relax
  \providecommand{\doi}[1]{doi: #1}\else
  \providecommand{\doi}{doi: \begingroup \urlstyle{rm}\Url}\fi

\bibitem[Chung et~al.(2024)Chung, Hou, Longpre, Zoph, Tay, Fedus, Li, Wang, Dehghani, Brahma, et~al.]{chung2024scaling}
Hyung~Won Chung, Le~Hou, Shayne Longpre, Barret Zoph, Yi~Tay, William Fedus, Yunxuan Li, Xuezhi Wang, Mostafa Dehghani, Siddhartha Brahma, et~al.
\newblock Scaling instruction-finetuned language models.
\newblock \emph{Journal of Machine Learning Research}, 25\penalty0 (70):\penalty0 1--53, 2024.

\bibitem[Deng et~al.(2022)Deng, Shiralkar, Lockard, Huang, and Sun]{deng2022dom}
Xiang Deng, Prashant Shiralkar, Colin Lockard, Binxuan Huang, and Huan Sun.
\newblock Dom-lm: Learning generalizable representations for html documents.
\newblock \emph{arXiv preprint arXiv:2201.10608}, 2022.

\bibitem[Deng et~al.(2023)Deng, Gu, Zheng, Chen, Stevens, Wang, Sun, and Su]{deng2023mindweb}
Xiang Deng, Yu~Gu, Boyuan Zheng, Shijie Chen, Samuel Stevens, Boshi Wang, Huan Sun, and Yu~Su.
\newblock Mind2web: Towards a generalist agent for the web.
\newblock In \emph{Thirty-seventh Conference on Neural Information Processing Systems Datasets and Benchmarks Track}, 2023.
\newblock URL \url{https://openreview.net/forum?id=kiYqbO3wqw}.

\bibitem[Deng et~al.(2024)Deng, Zhang, Zhang, Yuan, Ng, and Chua]{deng2024multi}
Yang Deng, Xuan Zhang, Wenxuan Zhang, Yifei Yuan, See-Kiong Ng, and Tat-Seng Chua.
\newblock On the multi-turn instruction following for conversational web agents.
\newblock \emph{arXiv preprint arXiv:2402.15057}, 2024.

\bibitem[Li et~al.(2023)Li, Zhang, Zhang, Long, Xie, and Zhang]{li2023towards}
Zehan Li, Xin Zhang, Yanzhao Zhang, Dingkun Long, Pengjun Xie, and Meishan Zhang.
\newblock Towards general text embeddings with multi-stage contrastive learning.
\newblock \emph{arXiv preprint arXiv:2308.03281}, 2023.

\bibitem[Liu et~al.(2018)Liu, Guu, Pasupat, and Liang]{zheran2018reinforcement}
Evan~Zheran Liu, Kelvin Guu, Panupong Pasupat, and Percy Liang.
\newblock Reinforcement learning on web interfaces using workflow-guided exploration.
\newblock In \emph{International Conference on Learning Representations}, 2018.
\newblock URL \url{https://openreview.net/forum?id=ryTp3f-0-}.

\bibitem[Lu et~al.(2024)Lu, Kasner, and Reddy]{pmlr-v235-lu24e}
Xing~Han Lu, Zden\v{e}k Kasner, and Siva Reddy.
\newblock {W}eb{LINX}: Real-world website navigation with multi-turn dialogue.
\newblock In Ruslan Salakhutdinov, Zico Kolter, Katherine Heller, Adrian Weller, Nuria Oliver, Jonathan Scarlett, and Felix Berkenkamp, editors, \emph{Proceedings of the 41st International Conference on Machine Learning}, volume 235 of \emph{Proceedings of Machine Learning Research}, pages 33007--33056. PMLR, 21--27 Jul 2024.
\newblock URL \url{https://proceedings.mlr.press/v235/lu24e.html}.

\bibitem[Popovi{\'c}(2015)]{popovic2015chrf}
Maja Popovi{\'c}.
\newblock chrf: character n-gram f-score for automatic mt evaluation.
\newblock In \emph{Proceedings of the tenth workshop on statistical machine translation}, pages 392--395, 2015.

\bibitem[Reimers and Gurevych(2019)]{reimers2019sentence}
Nils Reimers and Iryna Gurevych.
\newblock Sentence-bert: Sentence embeddings using siamese bert-networks.
\newblock \emph{arXiv preprint arXiv:1908.10084}, 2019.

\bibitem[Shi et~al.(2017)Shi, Karpathy, Fan, Hernandez, and Liang]{pmlr-v70-shi17a}
Tianlin Shi, Andrej Karpathy, Linxi Fan, Jonathan Hernandez, and Percy Liang.
\newblock World of bits: An open-domain platform for web-based agents.
\newblock In Doina Precup and Yee~Whye Teh, editors, \emph{Proceedings of the 34th International Conference on Machine Learning}, volume~70 of \emph{Proceedings of Machine Learning Research}, pages 3135--3144. PMLR, 06--11 Aug 2017.
\newblock URL \url{https://proceedings.mlr.press/v70/shi17a.html}.

\bibitem[Wang et~al.(2022)Wang, Fang, Ravula, Feng, Quan, and Liu]{wang2022webformer}
Qifan Wang, Yi~Fang, Anirudh Ravula, Fuli Feng, Xiaojun Quan, and Dongfang Liu.
\newblock Webformer: The web-page transformer for structure information extraction.
\newblock In \emph{Proceedings of the ACM Web Conference 2022}, pages 3124--3133, 2022.

\bibitem[Wang et~al.(2020)Wang, Wei, Dong, Bao, Yang, and Zhou]{wang2020minilm}
Wenhui Wang, Furu Wei, Li~Dong, Hangbo Bao, Nan Yang, and Ming Zhou.
\newblock Minilm: Deep self-attention distillation for task-agnostic compression of pre-trained transformers.
\newblock \emph{Advances in Neural Information Processing Systems}, 33:\penalty0 5776--5788, 2020.

\bibitem[Xiao et~al.(2023)Xiao, Liu, Zhang, and Muennighof]{xiao2023c}
Shitao Xiao, Zheng Liu, Peitian Zhang, and Niklas Muennighof.
\newblock C-pack: Packaged resources to advance general chinese embedding.
\newblock \emph{arXiv preprint arXiv:2309.07597}, 2023.

\bibitem[Yao et~al.(2022)Yao, Chen, Yang, and Narasimhan]{yao2022webshop}
Shunyu Yao, Howard Chen, John Yang, and Karthik Narasimhan.
\newblock Webshop: Towards scalable real-world web interaction with grounded language agents.
\newblock \emph{Advances in Neural Information Processing Systems}, 35:\penalty0 20744--20757, 2022.

\bibitem[Zheng et~al.(2024)Zheng, Gou, Kil, Sun, and Su]{zheng2024gptvision}
Boyuan Zheng, Boyu Gou, Jihyung Kil, Huan Sun, and Yu~Su.
\newblock {GPT}-4v(ision) is a generalist web agent, if grounded.
\newblock In \emph{Forty-first International Conference on Machine Learning}, 2024.
\newblock URL \url{https://openreview.net/forum?id=piecKJ2DlB}.

\bibitem[Zhou et~al.(2023)Zhou, Xu, Zhu, Zhou, Lo, Sridhar, Cheng, Ou, Bisk, Fried, Alon, and Neubig]{zhou2023webarena}
Shuyan Zhou, Frank~F. Xu, Hao Zhu, Xuhui Zhou, Robert Lo, Abishek Sridhar, Xianyi Cheng, Tianyue Ou, Yonatan Bisk, Daniel Fried, Uri Alon, and Graham Neubig.
\newblock Webarena: A realistic web environment for building autonomous agents.
\newblock In \emph{Second Agent Learning in Open-Endedness Workshop}, 2023.
\newblock URL \url{https://openreview.net/forum?id=rmiwIL98uQ}.

\bibitem[Zhou et~al.(2021)Zhou, Sheng, Vo, Edmonds, and Tata]{zhou2021simplified}
Yichao Zhou, Ying Sheng, Nguyen Vo, Nick Edmonds, and Sandeep Tata.
\newblock Simplified dom trees for transferable attribute extraction from the web.
\newblock \emph{arXiv preprint arXiv:2101.02415}, 2021.

\end{thebibliography}

\appendix

\section{Ethical Statement} \label{sec:Ethics}
Our work focuses on improving the generalization capabilities of multi-turn web navigation agents by optimizing state representation and context management. Our contributions aim to enhance the performance and applicability of these agents in real-world scenarios, but with these advancements come certain ethical considerations.

On the positive side, this work has the potential to greatly improve accessibility, particularly for individuals with disabilities, by making web navigation more intuitive and user-friendly. Enhanced context management and state representation can lead to more efficient digital assistants, automating repetitive and error-prone tasks and allowing knowledge workers to focus on more complex, creative, or high-level problem-solving activities. Additionally, the ability of these agents to better understand and align with user intent can foster more effective human-AI collaboration in various applications.

While necessary for the proposed training and inference processes (especially with the extended interaction history), collecting and processing large amounts of web navigation data could pose privacy risks. Data collection for applications of this research must adhere to the highest privacy standards and ensure that users are fully informed about how their data is being used. Furthermore, as these web navigation agents become more sophisticated, there exists a risk of misuse, such as automating fraudulent activities, spreading misinformation, or conducting unauthorized data scraping. 

We advocate for ongoing dialogue and collaboration within the AI research community to address these ethical challenges proactively and ensure that advancements in web navigation technology contribute positively to society.

\end{document}